\documentclass[sn-basic,Numbered]{sn-jnl}

\usepackage{graphicx}%
\usepackage{multirow}%
\usepackage{amsmath,amssymb,amsfonts}%
\usepackage{amsthm}%
\usepackage{mathrsfs}%
\usepackage[title]{appendix}%
\usepackage{xcolor}%
\usepackage{textcomp}%
\usepackage{manyfoot}%
\usepackage{booktabs}%
\usepackage{algorithm}%
\usepackage{algorithmicx}%
\usepackage{algpseudocode}%
\usepackage{listings}%
\usepackage{caption}
\usepackage{cleveref}
\usepackage{lscape}
\usepackage{comment}
\usepackage{arydshln}
\usepackage{appendix}
\usepackage{float}
\usepackage{subfigure}

\theoremstyle{thmstyleone}%
\theoremstyle{thmstyletwo}%

\theoremstyle{thmstylethree}%

\begin{document}

\title[Article Title]{Identifying Factors to Help Improve Existing Decomposition-Based PMI Estimation Methods}


\author*[1,2]{\fnm{Anna-Maria} \sur{Nau}}

\author[3]{\fnm{Phillip} \sur{Ditto}}

\author[3]{\fnm{Dawnie Wolfe} \sur{Steadman}}

\author[1]{\fnm{Audris} \sur{Mockus}}

\affil*[1]{\orgdiv{Department of Electrical Engineering and Computer Science}, \orgname{The University of Tennessee}, \orgaddress{\city{Knoxville}, \state{Tennessee}, \country{USA}}}

\affil[2]{\orgdiv{The Bredesen Center for Interdisciplinary Research and Graduate Education}, \orgname{The University of Tennessee}, \orgaddress{\city{Knoxville}, \state{Tennessee}, \country{USA}}}

\affil[3]{\orgdiv{Department of Anthropology}, \orgname{The University of Tennessee}, \orgaddress{\city{Knoxville}, \state{Tennessee}, \country{USA}}}


\abstract{Accurately assessing the postmortem interval (PMI), or the time since death is an important task in forensic science. Some of the existing techniques use regression models that use a decomposition score to predict the PMI or accumulated degree days (ADD), however, the provided formulas are based on very small samples and the accuracy is low. With the advent of Big Data, much larger samples can be used to improve PMI estimation methods. We, therefore, aim to investigate ways to improve PMI prediction accuracy by (a) using a much larger sample size, (b) employing more flexible and advanced linear models, and (c) enhancing models with demographic and environmental factors known to affect the human decay process. Specifically, this study involved the curation of a sample of 249 human subjects from a large-scale human decomposition photographic collection, followed by evaluating pre-existing PMI/ADD formulas and fitting increasingly sophisticated models to estimate the PMI/ADD. Results showed that including the total decomposition score (TDS), demographic factors (age, biological sex, and BMI), and weather-related factors (i.e., season of discovery, temperature history, and humidity history) increased the accuracy of the PMI/ADD models. Furthermore, the best performing PMI estimation model using the TDS, demographic, and weather-related features as predictors resulted in an adjusted R-squared of 0.34 and a RMSE of 0.95. It had a 7\% lower RMSE than a model using only the TDS to predict the PMI and a 48\% lower RMSE than the pre-existing PMI formula. The best ADD estimation model, also using the TDS, demographic, and weather-related features as predictors, resulted in an adjusted R-squared of 0.52 and a RMSE of 0.89. It had an 11\% lower RMSE than the model using only the TDS to predict the ADD and a 52\% lower RMSE than the pre-existing ADD formula. This work demonstrates the need (and way) to incorporate demographic and environmental factors into PMI/ADD estimation models.}




\maketitle

\section{Introduction}\label{introduction}
Determining the postmortem interval (PMI), or time since death, is an important task in human remains cases. An accurate estimation of the PMI can significantly narrow down the list of potential decedents the remains might correspond to, facilitating the eventual identification of the individual. In cases involving homicide, law enforcement professionals can utilize the PMI to eliminate potential suspects and validate witness accounts. Additionally, having knowledge of the PMI aids in establishing the spectrum of natural occurrences and environmental influences that impacted the remains as the seasons passed, thereby enabling a more comprehensive analysis.

Human decomposition is a natural and continuous process involving the breakdown of tissues after death that can span from weeks to years, contingent upon a range of biological and environmental factors. Comprehending how these factors influence the rate of human decomposition is pivotal in ascertaining the PMI, which is a crucial piece of information in forensic contexts and something that has long been sought after~\cite{megyesi,galloway1989decay,vass2011elusive}.

Over the years, researchers in the field of forensic anthropology have developed methods that divide the human decomposition process into high-level categories or stages of decay. Each stage is characterized by distinct decomposition phenomena, aiding in the estimation of the PMI~\cite{rodriguez1982insect,galloway1989decay}. One such method was developed by Gelderman et al.~\cite{gelderman}, a simplified version of Megyesi et al.'s \cite{megyesi} method, which involves a human decomposition scoring method and regression formulas to estimate the PMI and the accumulated degree days (ADD). The scoring method separates the human body into distinct anatomical regions, which are each assigned a score reflecting the amount of decomposition present. The individual scores are summed to obtain the total decomposition score (TDS), which is then used to estimate the PMI/ADD using their developed formulas.

Gelderman et al.'s \cite{gelderman} PMI/ADD estimation formulas only considered the TDS as a predictor variable, but what about other factors, such as donor demographics and environmental information? It is known from existing literature that a wide variety of factors, including body mass \cite{matuszewski2014effect,sutherland2013effect}, temperature \cite{johnson2013thermogenesis,vass2011elusive}, humidity \cite{vass2011elusive,cockle2015human}, and insect activity \cite{simmons2010debugging,payne1965summer} also affect the decomposition process and should be taken into consideration when determining the PMI. Therefore, expanding the TDS-only PMI/ADD formulas by including such factors could potentially improve their prediction accuracy.

Additionally, Gelderman et al.'s \cite{gelderman} PMI/ADD estimation formulas were developed and evaluated on a very small sample of human subjects. The sample size significantly affects the reliability, validity, and generalizability of study findings. In fact, it is a crucial factor that not only impacts the study's statistical power and precision but also influences its capacity to draw meaningful conclusions and apply the results to a wider context \cite{faber2014sample,sullivan2016common}.

The aim of this study is to investigate ways to improve decomposition-based PMI prediction by (a) using a much larger sample size, (b) employing more flexible and advanced linear models, and (c) enhancing the TDS-only models with demographic and environmental factors known to affect the human decay process. Specifically, various univariate and multivariate linear regression analyses using a combination of predictor variables to estimate the PMI and ADD will be conducted and evaluated. Additionally, Gelderman et al.'s \cite{gelderman} decomposition scoring method will be used to measure decomposition and calculate the TDS, and their PMI/ADD estimation formulas will be evaluated.

\section{Materials and methods}\label{materials_methods}
\subsection{The human decomposition photographic collection}\label{dataset}
The human decomposition photographic collection is a large-scale image dataset from which the study sample is obtained. The collection includes images of decomposing bodies donated to the Forensic Anthropology Center at The University of Tennessee, Knoxville. The center houses the Anthropology Research Facility (ARF), an outdoor decomposition laboratory. Forensic experts from the ARF captured these images at non-uniform intervals, with one or more days between each capture. The images, taken from various angles, depict different anatomical areas to illustrate the various stages and regions of human decomposition. The image resolutions vary from 2400 × 1600 up to 4900 × 3200. The dataset covers the period from 2011 to 2023 and comprises over 1.5 million images contributed by more than 800 donors. 

\subsection{The study sample}\label{study_sample}
\begin{figure}[ht]
    \centering
    \subfigure(a){\includegraphics[width=0.46\textwidth]{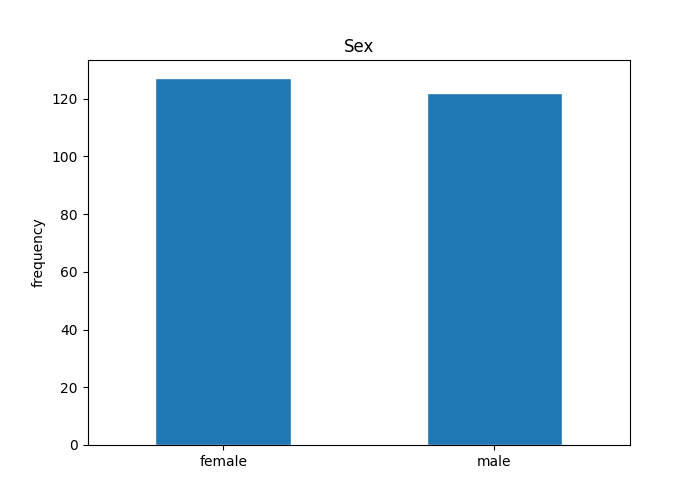}}
    \subfigure(b){\includegraphics[width=0.46\textwidth]{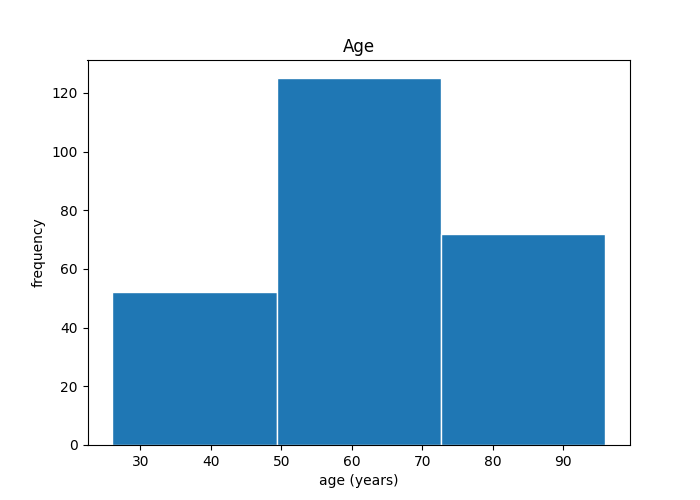}} 
    \subfigure(c){\includegraphics[width=0.46\textwidth]{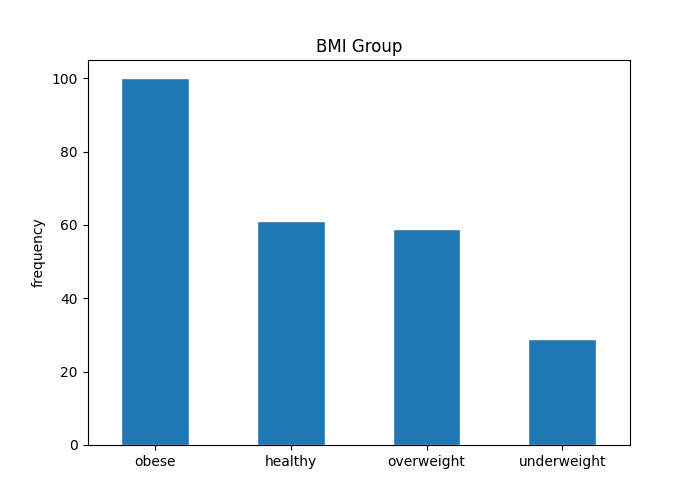}}
    \subfigure(d){\includegraphics[width=0.46\textwidth]{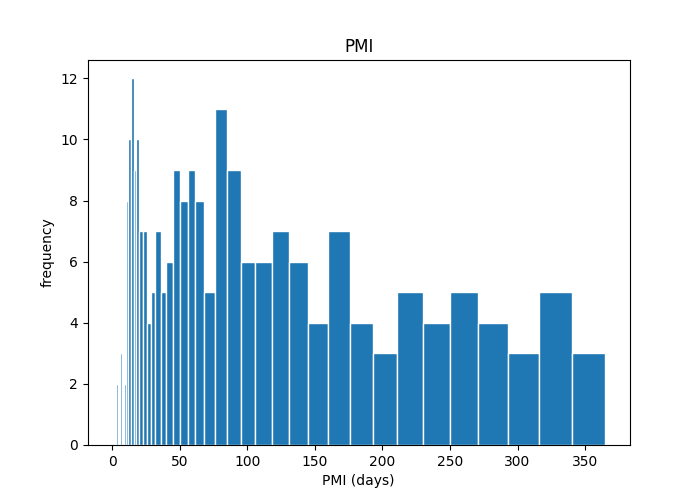}}
    \caption{Study sample (n=249) distribution plots by (a) sex, (b) age, (c) BMI group, and (d) PMI.}
    \label{fig:distribution_plots}
\end{figure}

The study sample was obtained from the large-scale human decomposition image dataset (see Section \ref{dataset}). Specifically, the subjects were chosen with respect to the following attributes: 
\begin{itemize}
  \item biological sex: female or male. 
  \item age groups: age $<$ 49 (younger), 49 $\le$ age $<$ 72 (middle), age $\ge$ 72 (older).
  \item body mass index (BMI) groups defined by the Center for Disease Control and Prevention (CDC): underweight (BMI $<$ 18.5), healthy (18.5 $\le$ BMI $<$ 25), overweight (25 $\le$ BMI $<$ 30), and obese (BMI $\ge$ 30).
  \item PMI less than 1 year (or 365 days).
\end{itemize}
By considering the above attributes, the objective was to create a study sample that was both diverse and balanced, meaning it included subjects across both sexes, different age and BMI groups, and PMIs. The final study sample consisted of 249 subjects, with the attribute distribution plots shown in Figure \ref{fig:distribution_plots}. Out of the 249 subjects, 127 were female and 122 were male, with ages ranging from 26 to 96 years. All subjects had a known PMI of less than one year.

All subjects were decomposing outdoors, unclothed, and complete, meaning no missing body parts so that the decomposition could be scored for the entire body in a standardized fashion. Additionally, no buried, burned, or submerged remains were used in this study because they decompose differently \cite{van2017aquatic}. Since all cases came from the ARF, they decomposed in a similar environmental setting - an open-wooded, mostly shaded area with the ground consisting of soil, gravel, and dead/decaying plant matter.

\subsection{Decomposition measurement}\label{scoring_method}
The state of decomposition was measured for each subject using the human decomposition scoring method proposed by Gelderman et al. \cite{gelderman} (see Appendix \ref{appendix_a}). To account for the differential decomposition that occurs in different body segments, this scoring method separates the human body into three anatomical regions: (1) the head (including the neck), (2) the torso, and (3) the limbs (including the hands and feet). Based on the morphological features present, each anatomical region is categorized into six stadia or stages, with the lowest (i.e., score 1) indicating no visible changes and the highest (i.e., score 6) indicating complete skeletonization. Once each anatomical region is scored, the three decomposition scores are summed to obtain the total decomposition score (TDS), which ranges from 3 to 18.

The decomposition scoring was performed by a forensic anthropologist who was well-trained in Gelderman et al.'s \cite{gelderman} scoring method. Specifically, the study sample photos (i.e., photos of the head/neck, torso, and limbs per subject) were assessed and scored on an in-house developed data visualization and annotation software called ICPUTRD (Image Cloud Platform for Use in Tagging and Research on Decomposition)~\cite{nau}. Once the decomposition scoring was completed, the TDS was calculated for each subject in the study sample.

\subsection{Weather data and ADD calculation}\label{weather_data}
Since the ADD in addition to the PMI will be estimated, temperature data was collected from to the closest National Weather Service station at McGhee Tyson Airport, Knoxville, TN. All temperature data (dry bulb) was in the form of daily averages and recorded in degrees Celsius (C). Following Gelderman et al. \cite{gelderman}, the ADD was calculated for each subject by adding the average daily temperatures above 0 degrees C (base temperature) from death until discovery.

\subsection{PMI/ADD estimation}\label{pmi_estimation}
For the PMI/ADD estimation the following were conducted: (1) evaluate Gelderman et al.'s \cite{gelderman} PMI/ADD formulas (\ref{eqn:pmi_base}) and (\ref{eqn:add_base}) on our larger study sample and (2) perform various univariate and multivariate linear regression analyses using different combinations of the following input variables to predict the PMI and ADD:
\begin{itemize}
  \item TDS
  \item Subject demographics, including biological sex, age at death, and BMI.
  \item Weather-related features, including season of discovery (winter, spring, summer, or fall), temperature history, and humidity history.
\end{itemize}

\begin{equation}
\label{eqn:pmi_base}
    \text{PMI} = 10^{(-0.93 + (0.18 \times \text{TDS}))}
\end{equation}

\begin{equation}
\label{eqn:add_base}
    \text{ADD} = 10^{(0.03 + (0.19 \times \text{TDS}))}
\end{equation}

In order to be used in the regression analysis, the categorical variables, sex and season of discovery, were converted to numeric by using one-hot-encoding. Specifically, one-hot-encoding creates a dummy variable for each category of the categorical variable with values 1 or 0 representing the presence or absence of the category. When including dummy variables in a regression analysis one needs to be aware of the the dummy variable trap, which is when two or more variables are highly correlated; in simple terms one variable can be predicted from the others. For instance, knowing male=0, implies female=1, or knowing spring=0, summer=0, fall=0, implies winter=1, and vice versa. The solution to the dummy variable trap is to drop one of the categorical variables, that is, if there are m number of categories, use m-1 in the regression analysis. In doing so, the categorical sex variable was converted to one dummy variable with values 1 for male and 0 for not male (hence female). Similarly, the categorical season of discovery variable was converted to numeric by creating three dummy variables (i.e., spring, summer, and fall), each with values 1 and 0. The BMI was calculated from the estimated cadaver weight (in pounds) and height (in inches). Similar to the ADD calculation (see \ref{weather_data}), to create the temperature and humidity history features, temperature (dry bulb) and humidity (relative) data were collected from the nearest National Weather Service station located at McGhee Tyson Airport, Knoxville, TN. Specifically, the temperature and humidity history features were calculated as follows for each subject in the study sample: (1) obtain daily average temperature and humidity values for the past two weeks from the date of discovery and (2) calculate the average for each two-week temperature and humidity history, resulting in the numeric variables called temp\_hist and hum\_hist, respectively.

\begin{figure}[h]
\centering
    \includegraphics[width=.7\columnwidth]{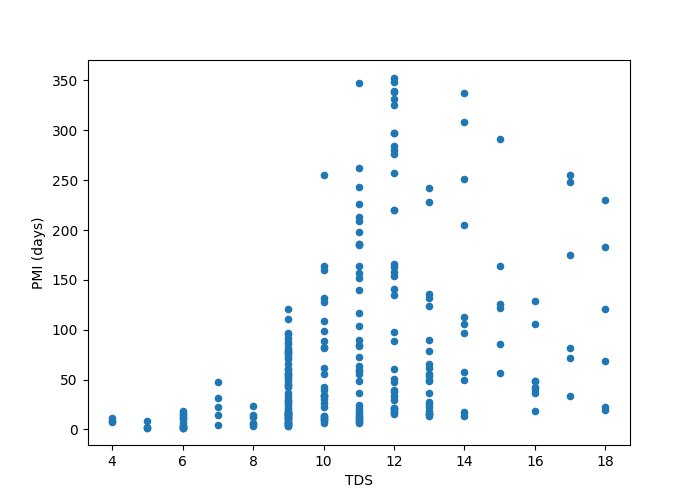}
    \caption{Plot of TDS vs. PMI (n=249).}
    \label{fig:TDS_PMI}
\end{figure}

\begin{figure}[h]
\centering
    \includegraphics[width=.7\columnwidth]{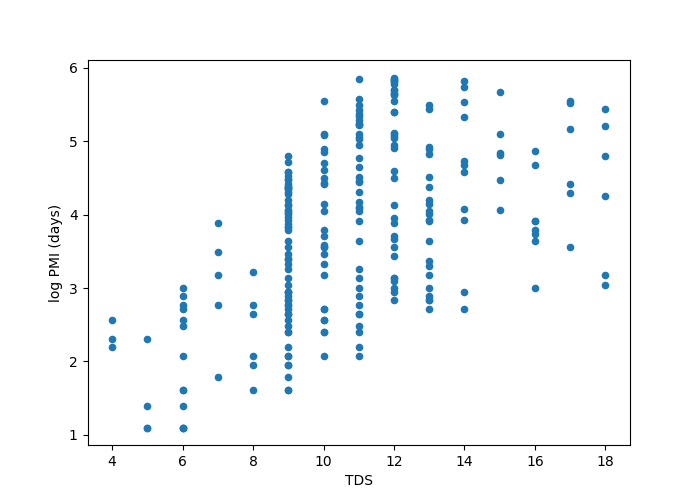}
    \caption{Plot of TDS vs. naturally log-transformed PMI (n=249).}
    \label{fig:TDS_logPMI}
\end{figure}

\begin{figure}[h]
\centering
    \includegraphics[width=.7\columnwidth]{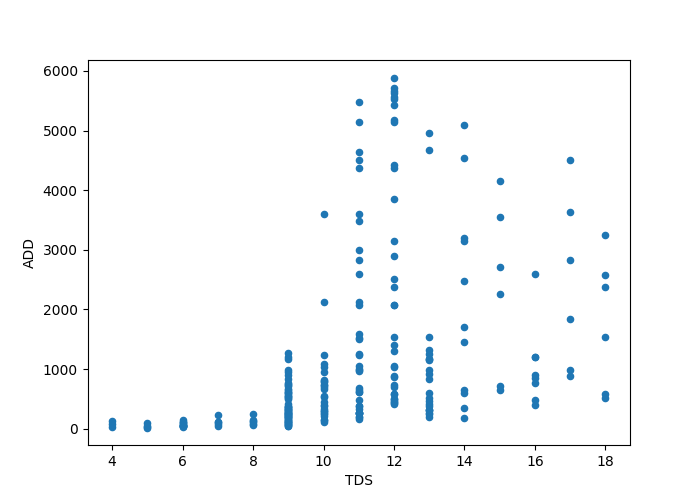}
    \caption{Plot of TDS vs. ADD (n=249).}
    \label{fig:TDS_ADD}
\end{figure}

\begin{figure}[h]
\centering
    \includegraphics[width=.7\columnwidth]{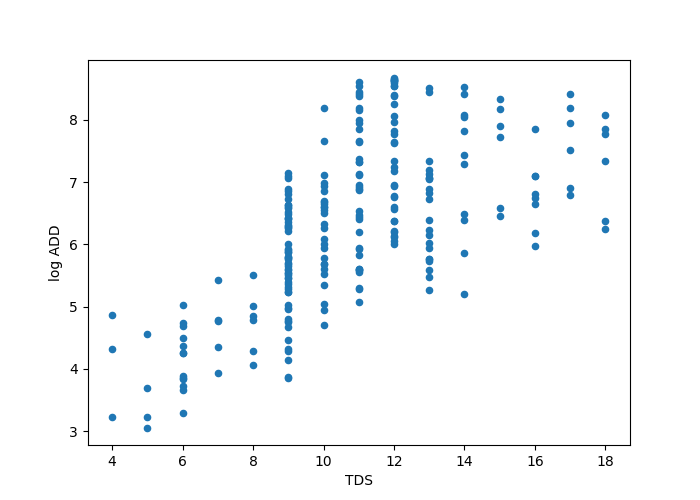}
    \caption{Plot of TDS vs. naturally log-transformed ADD (n=249).}
    \label{fig:TDS_logADD}
\end{figure}

The statistical analysis was conducted using SPSS Statistics (version 29.0.2.0) and the Python package, statsmodels. Specifically the following univariate and multivariate linear regression analysis using the study sample of 249 subjects were conducted and compared:
\begin{enumerate}
  \item TDS to predict the PMI and ADD (univariate).
  \item TDS + demographic features (sex, age, and BMI) to predict the PMI and ADD (multivariate).
  \item TDS + demographic features (sex, age, and BMI) + weather features (spring, summer, fall, temp\_hist and hum\_hist) to predict the PMI and ADD (multivariate).
\end{enumerate}
Similar to Gelderman et al. \cite{gelderman}, for the linear regression analysis, the PMI and ADD were naturally log-transformed to achieve a more linear relationship between the TDS and PMI, and the TDS and ADD as shown by \Cref{fig:TDS_PMI,fig:TDS_logPMI,fig:TDS_ADD,fig:TDS_logADD}. 

Additionally, all continuous independent variables, including the TDS, age, BMI, temp\_hist, and hum\_hist, were standardized to be on the same scale and thus contribute equally to the regression analysis. Z-score standardization was applied, rescaling the original variable to have a mean of zero and a standard deviation of one. Mathematically, this involves subtracting the mean of the original variable from the raw value and then dividing it by the standard deviation of the original variable, as shown in equation (\ref{eqn:standardization}).

\begin{equation}
    standardized \text{ } X =  \frac{X-mean}{standard \text{ } deviation}
    \label{eqn:standardization}
\end{equation}

The metrics used to evaluate the different regression analyses were the adjusted R-squared (adj. $R^2$) and root mean squared error (RMSE). The adj. $R^2$, as shown by equation (\ref{eqn:adj_r_sq}), is comparable to the $R^2$ or the coefficient of determination, as shown by equation (\ref{eqn:r_sq}), as its value lies between 0 and 1 (the higher, the better) and explains the proportion of variance for a dependent variable with respect to an independent variable(s) in the regression model. 

\begin{flalign} 
    R^2 &= 1 - \frac{RSS}{TSS}=1 - \frac{\sum (y_i - \hat{y_i})^2}{\sum (y_i - \bar{y})^2} \label{eqn:r_sq} \\
  \intertext{Where:}
    RSS &= \text{residual sum of squares}\notag\\
    TSS &= \text{total sum of squares}\notag\\
    y_i &= \text{actual value of sample $i$}\notag\\
    \hat{y_i} &= \text{predicted value of sample $i$}\notag\\
    \bar{y} &= \text{mean value of $y$}\notag
\end{flalign} 

\begin{flalign} 
    adj. R^2 &= \frac{(1-\text{$R^2$})(n-1)}{(n-k-1)} \label{eqn:adj_r_sq} \\
  \intertext{Where:}
    n &= \text{number of samples}\notag\\
    k &= \text{number of independent variables}\notag
\end{flalign} 

The difference between $R^2$ and adj. $R^2$ is that $R^2$ assumes that all the independent variables considered affect the result of the model, whereas the adj. $R^2$ considers only those independent variables which actually have an effect on the performance of the model. For instance, when multiple linear regression models are built, such as in this study with the forward addition method, at each iteration independent variables are added, the $R^2$ will keep increasing, but the adj. $R^2$ will only increase when the variable actually affects the dependent variable. If a variable is non-significant, the $R^2$ value will still increase, but the adj. $R^2$ value will decrease at that point. The RMSE, as shown by equation (\ref{eqn:rmse}), is the standard deviation of the residuals, measuring the average difference between a model's predicted values and the actual values. 

\begin{flalign} 
    RMSE &= \sqrt{\frac{\sum (y_i - \hat{y_i})^2}{n}} \label{eqn:rmse} \\
  \intertext{Where:}
    n &= \text{number of samples}\notag\\
    y_i &= \text{actual value of sample $i$}\notag\\
    \hat{y_i} &= \text{predicted value of sample $i$}\notag
\end{flalign} 

Residuals represent the distance between the regression line and the data points. As the data points move closer to the regression line, the model has less error, lowering the RMSE and producing more precise predictions. The RMSE values range from 0 (perfect predictions) to positive infinity and are in the same units as the dependent variable. Note, the RMSE is almost identical to the standard error of the regression. The difference is that the standard error adjusts for the number of independent variables while RMSE does not. Additionally, the regression coefficients and their p-values (with a significance level of .05) were calculated and reported. Since the predictor variables were standardized, the regression coefficients can be interpreted as the standardized regression coefficients. Finally, predicted vs. actual scatter plots will be created to visualize and compare the fit of the different regression analyses. In these plots, the x-axis represents the actual values, and the y-axis represents the predicted values. Ideally, if the predictions are perfect, the points will lie along a straight line with a slope of 1.


\section{Results}\label{results}
Table~\ref{table:twelve_months} shows the PMI/ADD estimation results. The first two rows (above the dashed line) give the results for Gelderman et al.'s~\cite{gelderman} PMI and ADD formulas evaluated on our study sample. To ensure equal comparison among all regression experiments, the log-transformed versions of Gelderman et al.'s~\cite{gelderman} PMI and ADD formulas were used, which required applying the base 10 logarithm to both sides of (\ref{eqn:pmi_base}) and (\ref{eqn:add_base}). This resulted in a RMSE of 1.84 when estimating the PMI and a RMSE of 1.86 when estimating the ADD. Note, the adj. $R^2$ was not calculated here since these formulas were evaluated and not fitted on our study sample. The remaining rows of Table \ref{table:twelve_months} give the results for the univariate and multivariate linear regression using different sets of input variables. For the univariate linear regression analysis using only the TDS as a predictor variable of PMI, an adj. $R^2$ of .26 and a RMSE of 1.02, and for ADD, an adj. $R^2$ of .41 and a RMSE of 1, were achieved. In both cases, the standardized coefficients were statistically significant with a p-value $<$ .05. For the multivariate linear regression analysis using the TDS and demographic features as predictor variables of PMI, an adj. $R^2$ of .27 and a RMSE of 1.01, and for ADD, an adj. $R^2$ of .42 and a RMSE of .99, were achieved. In both cases, the standardized coefficients for BMI had a p-value $<$ .05, hence statistically significant, and a p-value $>$ .05  for sex and age, hence not statistically significant. Lastly, for the multivariate linear regression analysis using the TDS, demographic, and weather features as predictor variables of PMI, an adj. $R^2$ of .34 and a RMSE of .95, and for ADD, an adj. $R^2$ of .52 and a RMSE of .89, were achieved. In both cases, the standardized coefficients of the weather features (i.e, spring, summer, fall, temp\_hist, and hum\_hist) had a p-value $<$ .05, hence statistically significant, and a p-value $>$ .05  for the demographic features (i.e., BMI, age, and sex), hence not statistically significant.

\setlength{\tabcolsep}{1pt} 
\renewcommand{\arraystretch}{1.5} 
\begin{sidewaystable*}
\centering
\begin{tabular}{ccccc}
Analysis (predictors, output) & adj. $R^2$ & RMSE &  Formula (back-transformed) \\
\hline\hline
    TDS, PMI (Gelderman et al. \cite{gelderman}) & - & 1.84 & See (\ref{eqn:pmi_base}) \\  
    TDS, ADD (Gelderman et al. \cite{gelderman}) & - & 1.86 & See (\ref{eqn:add_base}) \\  
\hdashline 
    TDS, PMI & .26 & 1.02 & PMI=$exp(1.48+0.21 TDS)$ \\
    \hline
    TDS, ADD & .41 & 1 & ADD=$exp(3.13+0.29 TDS)$ \\
    \hline
    TDS+demographics, PMI & .27 & 1.01 &  PMI=$exp(3.91 + 0.62 TDS + 0.1 BMI + 0.05 age - 0.22 sex)$\\
    \hline
    TDS+demographics, ADD & .42 & .99 & ADD=$exp(6.36 + 0.84 TDS + 0.14 BMI + 0.08 age - 0.06 sex)$\\
    \hline
    TDS+demographics+weather, PMI & .34 & .95 & $\begin{array}{c}
                \text{PMI} = exp(4.62 + 0.7 TDS + 0.06 BMI + 0.02 age \\
                - 0.16 sex - .42 spring - 1.3 summer - 1.13 fall\\ 
                + 0.37 temp\_hist - 0.15 hum\_hist) \\
                \end{array}$ \\
    \hline
    TDS+demographics+weather, ADD & .52 & .89 & $\begin{array}{c}
                \text{ADD} = exp(7.06 + 0.74 TDS + 0.07 BMI + 0.03 age \\
                - 0.01 sex - .53 spring - 1.27 summer - .93 fall\\ 
                + 0.76 temp\_hist - 0.15 hum\_hist)
                \end{array}$ \\
\hline\hline 
\end{tabular}
\caption{The univariate and multivariate linear regression analysis results. Reported are the adj. $R^2$, the RMSE, and the back-transformed (unlogged) formula with the standardized regression coefficients.}
\label{table:twelve_months}
\end{sidewaystable*}

\begin{figure}[!htb]
    \centering
    \subfigure(a){\includegraphics[width=0.46\textwidth]{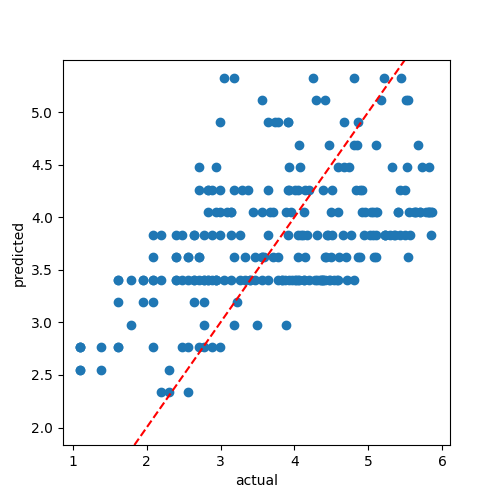}}
    \subfigure(b){\includegraphics[width=0.46\textwidth]{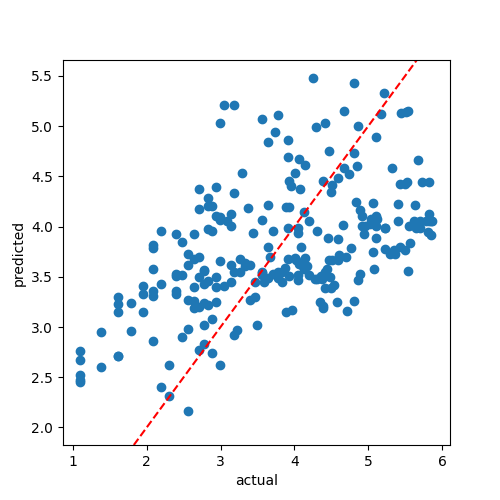}} 
    \subfigure(c){\includegraphics[width=0.46\textwidth]{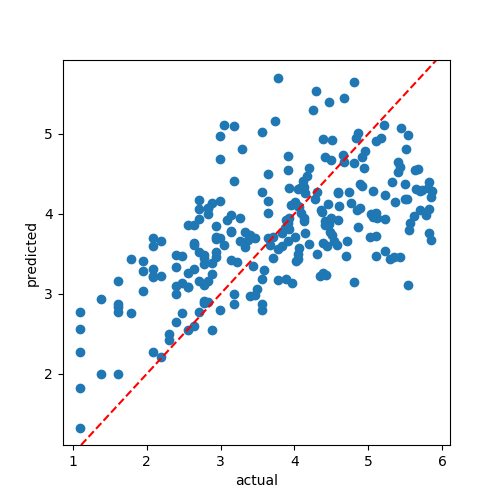}} 
    \caption{Predicted vs. actual PMI scatter plots for the different regression analyses: (a) TDS, (b) TDS + demographic features, (c) TDS + demographic features + weather features.}
    \label{fig:pred_actual_PMI}
\end{figure}

Figure \ref{fig:pred_actual_PMI} shows the predicted vs. actual PMI scatter plots for the regression analyses with input variables: (a) TDS, (b) TDS + demographic features, (c) TDS + demographic features + weather features. 

\begin{figure}[!htb]
    \centering
    \subfigure(a){\includegraphics[width=0.46\textwidth]{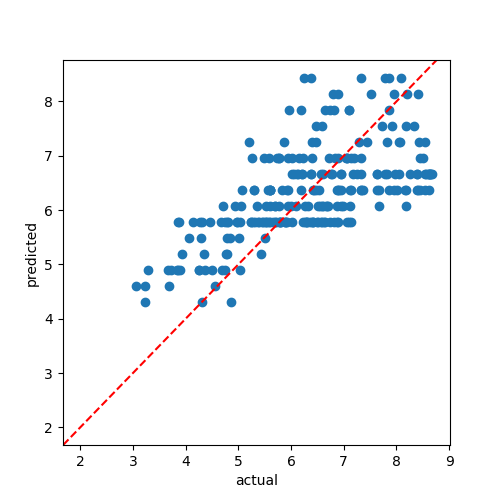}}
    \subfigure(b){\includegraphics[width=0.46\textwidth]{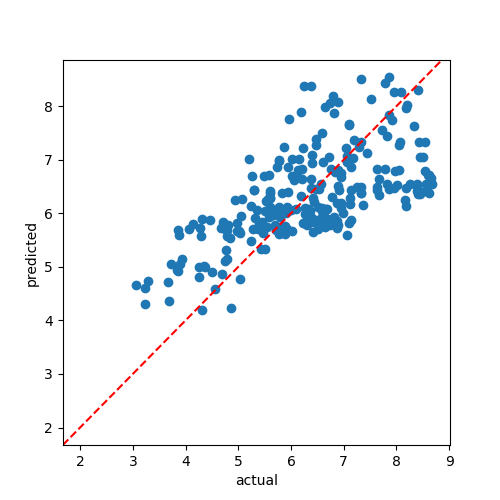}} 
    \subfigure(c){\includegraphics[width=0.46\textwidth]{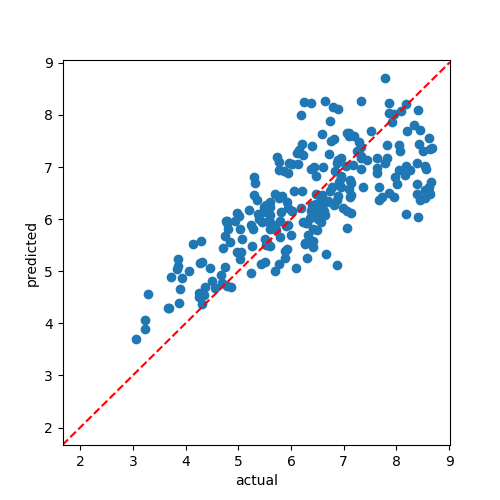}} 
    \caption{Predicted vs. actual ADD scatter plots for the different regression analyses with input variables: (a) TDS, (b) TDS + demographic features, (c) TDS + demographic features + weather features.}
    \label{fig:pred_actual_ADD}
\end{figure}

Similarly, Figure \ref{fig:pred_actual_ADD} shows the predicted vs. actual ADD scatter plots for the regression analyses with input variables with input variables: (a) TDS, (b) TDS + demographic features, (c) TDS + demographic features + weather features. The red dashed line gives the reference line x=y (i.e., actual=predicted) with slope 1.

\section{Discussion}\label{discussion}
The objective of this study was to investigate ways to improve decomposition-based PMI estimation from human remains by (a) using a much larger sample size, (b) employing more flexible and advanced linear models, and (c) enhancing PMI/ADD estimation models with additional factors, such as demographic and weather-related data. Specifically, pre-existing PMI/ADD formulas proposed by Gelderman et al.~\cite{gelderman} were evaluated on the significantly larger sample size. Additionally, various univariate and multivariate linear regression analyses using a variety of predictor variables to estimate the PMI and ADD were conducted and evaluated.

When evaluating Gelderman et al.'s \cite{gelderman} outdoor PMI/ADD formulas on our study sample, the results were poor. This is not surprising given the small sample size (i.e., 12 outdoor cases compared to the 249 outdoor cases of this study) the formulas were created with, resulting in low generalizability. In fact, Gelderman et al.'s \cite{gelderman} PMI formula resulted in a RMSE that was 48\% higher than the RMSE of the PMI model that considered the TDS, demographic, and weather-related features as predictor variables. Similarly, Gelderman et al.'s \cite{gelderman} ADD formula resulted in a RMSE that was 52\% higher than the RMSE of the ADD model that considered the TDS, demographic, and weather-related features as predictor variables. 

The univariate and multivariate linear regression analysis results showed that including additional factors, such as demographic and weather-related information, in addition to the TDS, improves PMI/ADD prediction performance. When estimating the PMI, the adjusted $R^2$ was 31\% higher and the RMSE was 7\% lower when using the TDS, demographic, and weather-related features compared to only using the TDS as predictor variables. Similarly, when estimating the ADD, the adjusted $R^2$ was 27\% higher and the RMSE was 11\% lower when using the TDS, demographic, and weather-related features compared to only using the TDS as predictor variables. This increase in prediction performance is also seen in the predicted vs. actual scatter plots (see Figures \ref{fig:pred_actual_PMI} and \ref{fig:pred_actual_ADD}) - as more predictor variables are considered, the points are closer to the red-dashed line (reference line), indicating a better fit and more accurate predictions. When assessing variable importance, the standardized coefficient of the TDS was statistically significant for the TDS-only PMI/ADD models. When considering the demographic features (i.e., age, sex, and BMI) in addition to the TDS, the standardized coefficients for TDS and BMI were statistically significant, but not for the age and sex variables when predicting the PMI and ADD. Lastly, when also considering the weather-related features (i.e., season of discovery, temperature history, and humidity history), the TDS and all weather-related features had statistically significant standardized coefficients, but not the age, sex, and BMI. This suggests that when using the TDS in combination with weather-related features such as the ones considered in this study, the demographic features may not have a significant effect on the dependent variables, PMI and ADD, and could thus be removed from the model. This will be further explored and verified in future work. 

While the methods described in this research build upon and provide improvements to previous methods, there are some important limitations of this work. For one,
the decomposition measurement to obtain the TBS for each subject in the study sample was conducted by a single forensic expert. This may introduce labeling bias and errors, in particular for large sample sizes, resulting in less accurate models~\cite{jiang2020identifying}. Therefore, future work will focus on creating a so-called ``gold standard" TDS dataset—a dataset carefully labeled and evaluated by multiple forensic experts. Such a dataset would be accepted as the most accurate and reliable of its kind. This step will ensure that the models are created with accurate and unbiased data, which is vital for developing high-quality models.

An additional limitation is that this study is environmental- and climate-specific. All study sample subjects decayed outdoors in an open-wooded area with the ground consisting of soil, gravel, and dead/decaying plant matter (e.g., rotting wood and shedding leaves). Additionally, the climate of this area is humid subtropical, characterized by high summer and moderate winter temperatures. Therefore, the PMI/ADD formulas of this study may not perform as well for different climate conditions or environments.

And finally, the PMI/ADD formulas reported in this study should be used with caution. The aim of this work was not to create new PMI/ADD estimation formulas but to investigate ways to improve decomposition-based PMI estimation methods, such as the one proposed by Gelderman et al.~\cite{gelderman}, by using a much larger sample size and including different factors known to affect the human decay process. Researchers are encouraged to evaluate and build upon our method to further assess its accuracy.

\section{Conclusion}\label{conclusion}
In conclusion, this study underscores the significance of leveraging larger datasets and incorporating a broader range of predictive factors to enhance PMI/ADD estimation models in forensic science. By integrating demographic and environmental variables with the TDS, the accuracy of PMI/ADD predictions compared to traditional models were significantly improved. The results suggest that a more comprehensive approach, which accounts for factors such as age, biological sex, BMI, season, temperature, and humidity, can better capture the complexity of human decomposition. This advancement not only highlights the potential for more accurate postmortem interval assessments but also lays the groundwork for future research to further refine and validate these models.

\backmatter


\bibliography{references}


\begin{appendices}
\section{Gelderman et al.'s \cite{gelderman} decomposition scoring method}\label{appendix_a}

\setlength{\tabcolsep}{2pt} 
\renewcommand{\arraystretch}{1} 
\begin{sidewaystable}[h!]
\centering
\begin{tabular}{llp{12cm}} 
Region & Score & Description \\
\hline\hline
    \multirow{11}{*}{Head} & 1 & 1.1 No visible changes\\
         & \multirow[t]{3}{*}{2} & 2.1 Livor mortis, rigour mortis and vibices \\
         &                    & 2.2 Eyes: cloudy and/or tache noir \\
         &                    & 2.3 Discoloration: brownish shades particularly at the edges. Drying of nose, ears and lips \\
         & \multirow[t]{3}{*}{3} & 3.1 Grey to green discoloration \\
         &                    & 3.2 Bloating of neck and face is present and/or skin blisters, skin slippage and/or marbling \\
         &                    & 3.3 Purging of decompositional fluids out of ears, nose and mouth and/or brown to black discoloration \\
         & \multirow[t]{2}{*}{4} & 4.1 Caving in of the flesh and tissues of eyes and throat. Skin having a leathery appearance \\
         &                    & 4.2 Partial skeletonization, joints still together \\       
         & 5 & 5.1 Gross skeletonization, some joints disarticulated\\
         & 6 & 6.1 Complete skeletonization\\
\hline
    \multirow{11}{*}{Torso} & 1 & 1.1 No visible changes \\
         & 2 & 2.1 Livor mortis, rigour mortis and vibices \\
         & \multirow[t]{4}{*}{3} & 3.1 Grey to green discoloration \\
         &                    & 3.2 Bloating with green discoloration and/or skin blisters, skin slippage and/or marbling \\
         &                    & 3.3 Rectal purging of decompositional fluids \\
         &                    & 3.4 Post-bloating: release of abdominal gasses with discoloration changing from green to black \\
         & \multirow[t]{3}{*}{4} & 4.1 Decomposition of tissue producing sagging of flesh. Caving in of the abdominal cavity \\
         &                    & 4.2 Skin having a leathery appearance \\   
         &                    & Partial skeletonization, joints still together \\   
         & 5 & 5.1 Gross skeletonization, some joints disarticulated\\
         & 6 & 6.1 Complete skeletonization\\
\hline
    \multirow{10}{*}{Limbs} & 1 & 1.1 No visible changes \\
         & \multirow[t]{2}{*}{2} & 2.1 2.1 Livor mortis, rigour mortis and vibices \\
         &                    & 2.2 Discoloration: brownish shades particularly at the edges. Drying of fingers and toes \\ 
         & \multirow[t]{3}{*}{3} & 3.1 Skin blisters and/or skin slippage and/or marbling \\
         &                    & 3.2 Grey to green discoloration \\
         &                    & 3.3 Brown to black discoloration \\
         & \multirow[t]{2}{*}{4} & 4.1 Skin having a leathery appearance \\
         &                    & 4.2 Partial skeletonization, joints and tendons still together \\ 
         & 5 & 5.1 Gross skeletonization, some joints disarticulated\\
         & 6 & 6.1 Complete skeletonization\\
\hline\hline 
\end{tabular}
\caption{Gelderman et al.'s \cite{gelderman} decomposition scoring method used in this study. Point.1, point.2, etc. represent the different phenomena.}
\label{app:scoring_method}
\end{sidewaystable}

\end{appendices}

\end{document}